\documentclass[10pt,twocolumn,letterpaper]{article}

\usepackage{cvpr}              

\usepackage{graphicx}
\usepackage{amsmath}
\usepackage{amssymb}
\usepackage{booktabs}

\usepackage{mathtools}
\usepackage{boldline}
\usepackage[dvipsnames]{xcolor}
\usepackage[normalem]{ulem}
\usepackage{enumitem}
\usepackage{tabularx}
\usepackage[ruled,vlined]{algorithm2e}

\usepackage[pagebackref,breaklinks,colorlinks]{hyperref}

\usepackage[capitalize]{cleveref}
\crefname{section}{Sec.}{Secs.}
\Crefname{section}{Section}{Sections}
\Crefname{table}{Table}{Tables}
\crefname{table}{Tab.}{Tabs.}

\def\cvprPaperID{7104} 
\def\confName{CVPR}
\def\confYear{2022}

\begin{document}

\title{Estimating Egocentric 3D Human Pose in the Wild with External Weak Supervision}


\author{Jian Wang\textsuperscript{1,2}~~~~~~Lingjie Liu\textsuperscript{1,2}~~~~~~Weipeng Xu\textsuperscript{3}~~~~~~Kripasindhu Sarkar\textsuperscript{1,2}\\
	Diogo Luvizon\textsuperscript{1,2}~~~~~~Christian Theobalt\textsuperscript{1,2}\\
	\textsuperscript{1}MPI Informatics~~~~~\textsuperscript{2}Saarland Informatics Campus~~~~~\textsuperscript{3}Facebook Reality Labs\\
	{\tt\small \{jianwang,lliu,ksarkar,theobalt\}@mpi-inf.mpg.de~~xuweipeng@fb.com}
}
\maketitle

\begin{abstract}
Egocentric 3D human pose estimation with a single fisheye camera has drawn a significant amount of attention recently. However, existing methods struggle with pose estimation from in-the-wild images, because they can only be trained on synthetic data due to the unavailability of large-scale in-the-wild egocentric datasets. Furthermore, these methods easily fail when the body parts are occluded by or interacting with the surrounding scene. To address the shortage of in-the-wild data, we collect a large-scale in-the-wild egocentric dataset called \emph{Egocentric Poses in the Wild (EgoPW)}. This dataset is captured by a head-mounted fisheye camera and an auxiliary external camera, which provides an additional observation of the human body from a third-person perspective during training.
We present a new egocentric pose estimation method, which can be trained on the new dataset with weak external supervision. 
Specifically, we first generate pseudo labels for the EgoPW dataset with a spatio-temporal optimization method by incorporating the external-view supervision.
The pseudo labels are then used to train an egocentric pose estimation network. To facilitate the network training, we propose a novel learning strategy to supervise the egocentric features with the high-quality features extracted by a pretrained external-view pose estimation model. 
The experiments show that our method predicts accurate 3D poses from a single in-the-wild egocentric image and outperforms the state-of-the-art methods both quantitatively and qualitatively. 
\end{abstract}

\section{Introduction}


Egocentric motion capture using head- or body-mounted cameras has recently become popular because traditional motion capture systems with outside-in cameras have limitations when the person is moving around in a large space and thus restrict the scope of applications. Different from traditional systems, the egocentric motion capture system is mobile, flexible, and has no requirements on recording space, which enables capturing a wide range of human activities for many applications, such as wearable medical monitoring, sports analysis, and $x$R.  

\begin{figure}
\begin{center}
\includegraphics[width=0.97\linewidth]{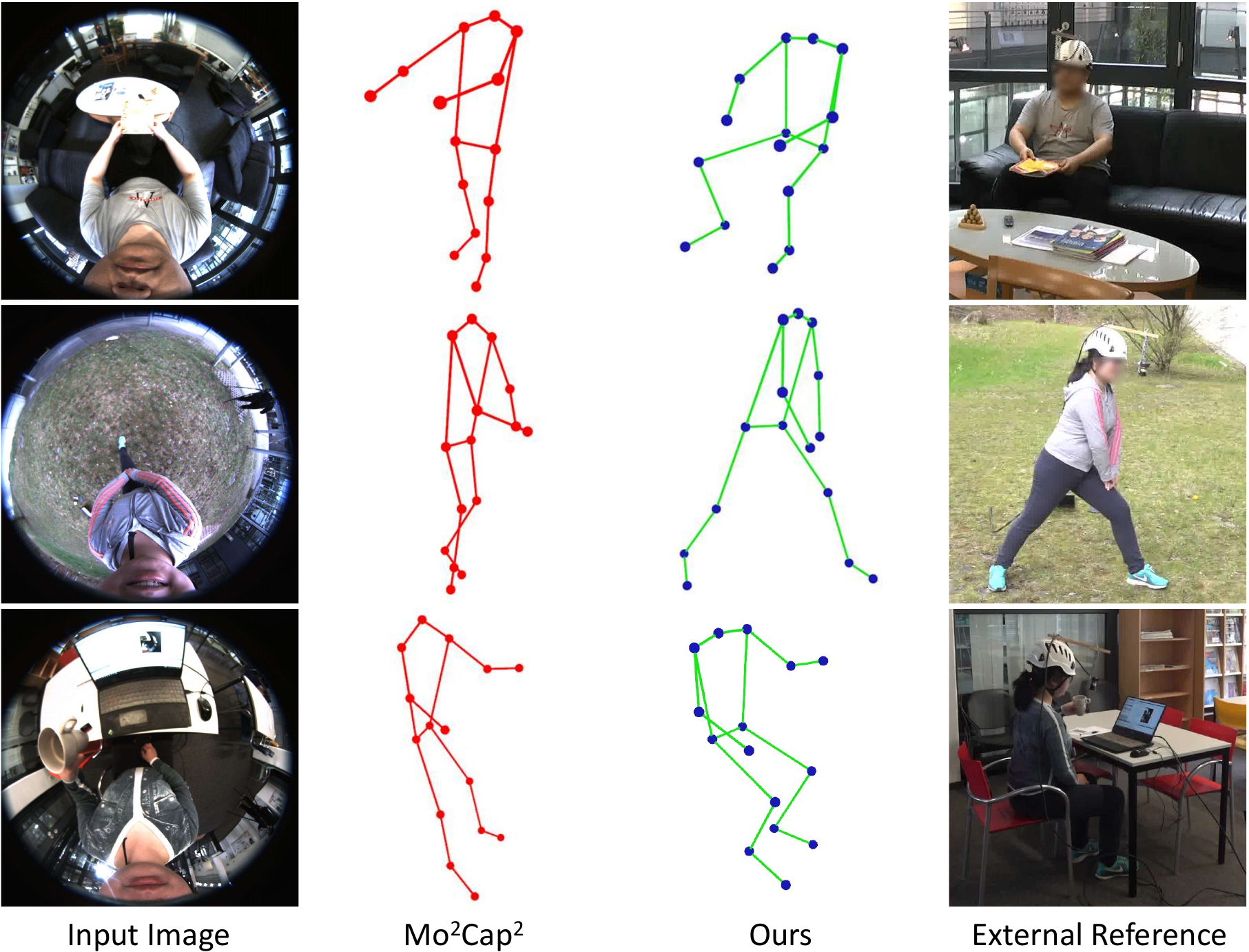}
\end{center}
   \caption{Compared with Mo$^2$Cap$^2$, our method gets a more accurate egocentric pose from a single in-the-wild image, especially when the body parts are occluded. Note that the external images are only used for visualization, not the inputs to our method.}
\label{fig:teaser}
\end{figure}

In this work, we focus on estimating the full 3D body pose from a single head-mounted fisheye camera. The most related works are Mo$^2$Cap$^2$~\cite{DBLP:journals/tvcg/XuCZRFST19} and $x$R-egopose~\cite{DBLP:conf/iccv/TomePAB19}. 
While these methods have produced compelling results, they are only trained on synthetic images as limited real data exists and, therefore, suffer from significant performance drop on real-world scenarios.
Furthermore, these methods often struggle with the cases when parts of the human body are occluded by or interacting with the surrounding scene (see the Mo$^2$Cap$^2$ results in Fig.~\ref{fig:teaser}).
This is due to the domain gap between synthetic and real data, but also due to their limited capability of handling occlusions.

To address the issue of the limited real egocentric data, we capture a large-scale in-the-wild egocentric dataset called \emph{Egocentric Poses in the Wild (EgoPW)}. This is currently the largest egocentric in-the-wild dataset, containing more than 312k frames and covering 20 different daily activities in 8 everyday scenes. To obtain the supervision for the network training, one possibility is using a multi-view camera setup to capture training data with ground truth 3D body poses or apply multi-view weak supervision.
%
%
However, this setup is impractical for recording in an environment with limited space (e.g. in the small kitchen shown in Fig.~\ref{fig:optimization_example}), which is a common recording scenario. Therefore, considering a trade-off between flexibility and 3D accuracy, we use a new device setup consisting of an egocentric camera and a single auxiliary external camera. We demonstrate that the external view can provide additional supervision during training, especially for the highly occluded regions in the egocentric view (e.g. the lower body part). 

To handle occlusions and estimate accurate poses, we propose a new egocentric pose estimation method for training on the EgoPW dataset in a weakly supervised way.
Specifically, we propose a spatio-temporal optimization method to generate accurate 3D poses for each frame in the EgoPW dataset. The generated poses are further used as pseudo labels for training an egocentric pose estimation network~\cite{DBLP:journals/tvcg/XuCZRFST19}. To improve the network performance, we facilitate the training of the egocentric pose estimation network with the extracted features from the external pose estimation network which has been trained on a large in-the-wild body pose dataset. 
Specifically, we enforce the feature extracted from these two views to be similar by fooling a discriminator not being able to detect which view the features are from. 
To further improve the performance of the pose estimation network, besides the EgoPW dataset, we also use a synthetic dataset~\cite{DBLP:journals/tvcg/XuCZRFST19} to train the network and adopt a domain adaptation strategy to minimize the domain gap between synthetic and real data.

We evaluate our method on the test data provided by Wang~\etal~\cite{wang2021estimating} and Xu~\etal~\cite{DBLP:journals/tvcg/XuCZRFST19}. Our method significantly outperforms the state-of-the-art methods both quantitatively and qualitatively. We also show qualitative results on various in-the-wild images, demonstrating that our method can predict accurate 3D poses on very challenging scenes, especially when the body joints are seriously occluded (see our results in Fig.~\ref{fig:teaser}). To summarize, our contributions are presented as follows:

\begin{itemize}
\itemsep -0.1em
    \item A new method to estimate egocentric human pose with weak supervision from an external view, which significantly outperforms existing methods on in-the-wild data, especially when severe occlusions exist;
    \item A large in-the-wild egocentric dataset (EgoPW) captured with a head-mounted fisheye camera and an external camera. We will make it publicly available;
    \item A new optimization method to generating pseudo labels for the in-the-wild egocentric dataset by incorporating the supervision from an external view; 
    \item  An adversarial method for training the network by learning the feature representation of egocentric images with external feature representation. 
\end{itemize}

\section{Related Work}

\paragraph{Egocentric 3D full body pose estimation.}
Rhodin \etal~\cite{DBLP:journals/tog/RhodinRCISSST16} developed the first method to estimate the full-body pose from a helmet-mounted stereo fisheye camera. Cha~\etal~\cite{cha2018fullycapture} presented an RNN-based method to estimate body pose with two pinhole cameras mounted on the head. Xu~\etal~\cite{DBLP:journals/tvcg/XuCZRFST19} introduced a single wide-view fisheye camera setup and proposed a single-frame based egocentric motion capture system. With the same setup, Tome \etal~\cite{DBLP:conf/iccv/TomePAB19} captured the egocentric pose with an auto-encoder network which captures the uncertainty in the predicted heatmaps. In order to further mitigate the effect of image distortions, Zhang~\etal~\cite{zhang2021automatic} proposed an automatic calibration module.  Hwang \etal \cite{Hwang2020} put an ultra-wide fisheye camera on the user's chest and estimate body pose, camera rotation and head pose simultaneously.
Jiang~\etal~\cite{jiang2021egocentric} mounted a front-looking fisheye camera on the user's head and estimated the body and head pose by leveraging the motion of the environment and extremity of the human body.
Wang~\etal~\cite{wang2021estimating} proposed an optimization algorithm to obtain temporally stable egocentric poses with motion prior learned from Mocap datasets.
However, these methods are all trained on synthetic datasets, thus suffering from the performance drop on the real images due to the domain gap and lack of external supervision. Our method, on the contrary, achieves better performance on the in-the-wild scenes. 

\begin{figure*}
\centering
\includegraphics[width=0.96\linewidth]{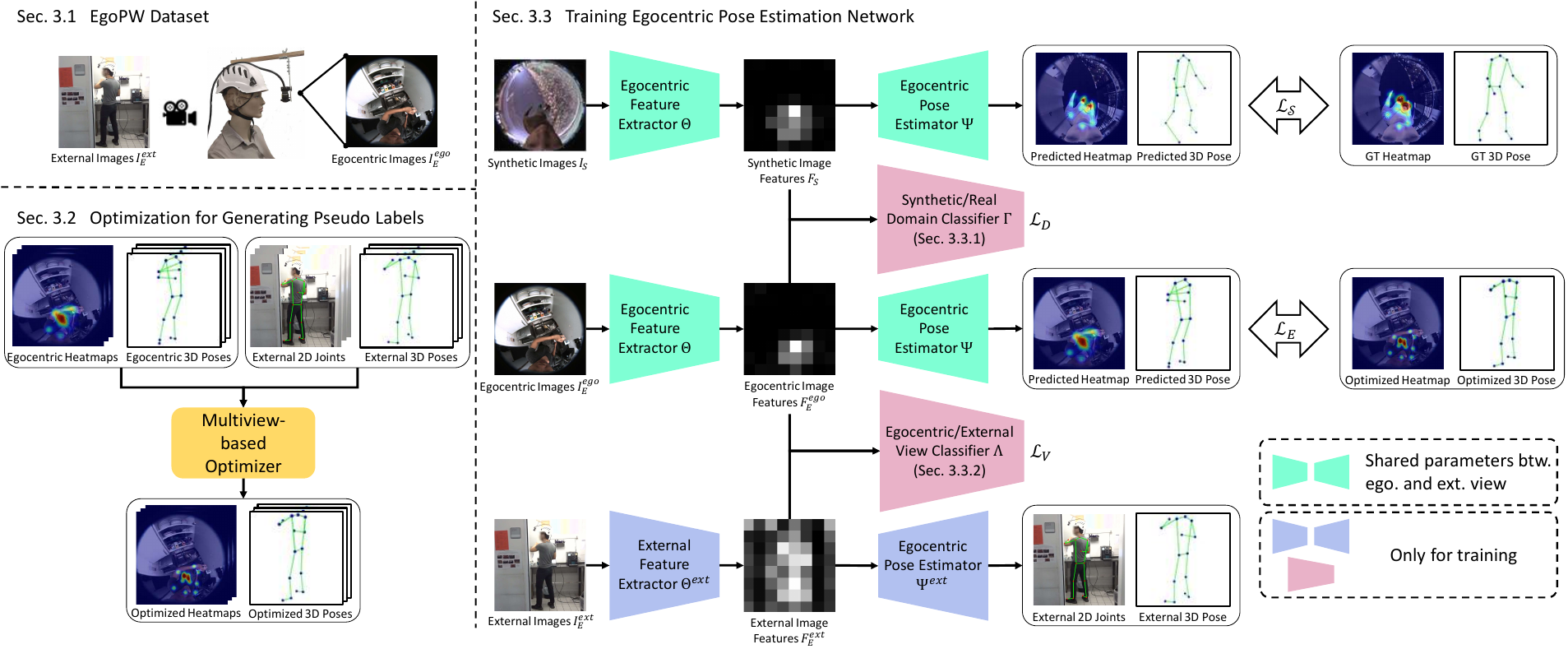}
\caption{Overview of our method.
We first collected the new EgoPW dataset (Sec.~\ref{subsec:dataset}), where pseudo labels are generated by a multi-view based optimization method (Sec.~\ref{subsec:optimizer}). We then train our model with the proposed framework (Sec.~\ref{subsec:train}), where the network is simultaneously trained with EgoPW datasets and synthetic data from Mo$^2$Cap$^2$. We enforce the egocentric network to learn a better feature representation from the external view (Sec.~\ref{subsubsec:ext_adv}) and bridge the gap between synthetic and real data with a domain classifier (Sec.~\ref{subsubsec:da}).
%
}
\label{fig:framework}
\end{figure*}



\paragraph{Pseudo label generation.}
The task of pseudo-labeling~\cite{lee2013pseudo, shi2018transductive,yang2019pseudo} is a semi-supervised learning technique that generates pseudo labels for unlabeled data and uses the generated labels to train a new model. This has been applied in the areas of segmentation~\cite{li2020self,zou2020pseudoseg,zheng2021rectifying,chen2020digging}, pose estimation~\cite{cao2019cross,li2021synthetic,mu2020learning,ludwig2021self} and image classification~\cite{hu2021simple,pham2021meta,arazo2020pseudo}.
Since the pseudo labels may be inaccurate, some methods have been proposed to filter inaccurate labels to increase the labeling stability. Shi~\etal~\cite{shi2018transductive} set confidence levels on unlabeled samples by measuring the sample density. Chen~\etal~\cite{chen2019progressive} enforced the stability of pseudo labels by adopting an easy-to-hard transfer strategy. Wang and Wu~\cite{wang2020repetitive} introduced a repetitive re-prediction strategy to update the pseudo labels, while Rizve~\etal~\cite{rizve2021defense} proposed an uncertainty-aware pseudo-label selection framework that selects pseudo labels. Morerio~\etal~\cite{morerio2020generative} used a conditional GAN to filter the noise in the pseudo labels.
Different from previous pseudo-labeling works which generate the labels from network predictions or clustering, we design an optimization framework to generate labels with supervision from egocentric and external views simultaneously. 

\paragraph{Weakly Supervised 3D Human Pose Estimation.}
Recently, there is a growing interest in developing weakly-supervised 3D pose estimation methods. Weakly-supervised methods do not require datasets with paired images and 3D annotations. Some works~\cite{wang2019distill,novotny2019c3dpo} leverages the non-rigid SFM to get 3D joint positions from 2D keypoint annotations in unconstrained images. Some works~\cite{drover2018can,chen2019unsupervised,wandt2019repnet,pavllo20193d,chen2019weakly} present an unsupervised learning approach to train the 3D pose estimation network with the supervision from 2D reprojections. The closest to our work are the approaches of~\cite{iqbal2020weakly,rhodin2018learning,wandt2021canonpose,kocabas2019self} in that they leverage the weak supervision from multi-view images for training. 
Iqbal~\etal~\cite{iqbal2020weakly} and Rhodin~\etal~\cite{rhodin2018learning} supervise the network training process by calculating the differences between Procrustes-aligned 3D poses from different views. Wandt~\etal~\cite{wandt2021canonpose} predict the camera poses and 3D body poses in a canonical form, and then supervise the training with the multi-view consistency. Kocabas~\etal~\cite{kocabas2019self} obtain the pseudo labels with epipolar geometry between different views and use the pseudo labels to train the 3D pose lifting network.
Different from previous works~\cite{iqbal2020weakly,rhodin2018learning,wandt2021canonpose,kocabas2019self}, our method uses spatio-temporal optimization framework that takes egocentric and external view as input to obtain robust 3D pseudo labels for training the network. This optimization method ensures the stability of the network training process when the 2D pose estimations are inaccurate.

\section{Method}
We propose a new approach to train a neural network on the in-the-wild dataset with weak supervision from egocentric and external views.
The overview of our approach is illustrated in Fig.~\ref{fig:framework}. We first capture a large-scale egocentric in-the-wild dataset, called \emph{EgoPW}, which contains synchronized egocentric and external image sequences (Sec.~\ref{subsec:dataset}). Next, we generate pseudo labels for the EgoPW dataset with an optimization-based framework. This framework takes as input a sequence in a time window with $B$ frames of egocentric images $\mathcal{I}^{ego}_{seq}=\{\mathcal{I}^{ego}_{1},\dots,\mathcal{I}^{ego}_{B}\}$ and external images $\mathcal{I}^{ext}_{seq}=\{\mathcal{I}^{ext}_{1},\dots,\mathcal{I}^{ext}_{B}\}$ and outputs egocentric 3D poses $\mathcal{P}^{ego}_{seq}=\{\mathcal{P}^{ego}_{1},\dots,\mathcal{P}^{ego}_{B}\}$ as the pseudo labels (Sec.~\ref{subsec:optimizer}). 
Next, we train the egocentric pose estimation network on the synthetic data from Mo$^2$Cap$^2$~\cite{DBLP:journals/tvcg/XuCZRFST19}
and on the EgoPW dataset with pseudo labels $\mathcal{P}^{ego}_{seq}$.
In the training process, we leverage the feature representation from an on-the-shelf external pose estimation network~\cite{xiao2018simple} to enforce our egocentric network to learn a better feature representation
in an adversarial way (Sec.~\ref{subsubsec:ext_adv}). We also use an adversarial domain adaptation strategy to mitigate the domain gap between synthetic and real datasets (Sec.~\ref{subsubsec:da}).

\subsection{EgoPW Dataset}\label{subsec:dataset}

We first describe the newly collected \emph{EgoPW} dataset, which is the first large-scale in-the-wild human performance dataset captured by an egocentric camera and an external camera (Sony RX0), both synchronized.
EgoPW contains a total of 318k frames, which are divided into 97 sequences of 10 actors in 20 clothing styles performing 20 different actions, including
\textit{reading magazine/newspaper},
\textit{playing board games},
\textit{doing a presentation},
\textit{walking},
\textit{sitting down},
\textit{using a computer},
\textit{calling on the phone},
\textit{drinking water},
\textit{writing on the paper},
\textit{writing on the whiteboard},
\textit{making tea},
\textit{cutting vegetables},
\textit{stretching},
\textit{running},
\textit{playing table tennis},
\textit{playing baseball},
\textit{climbing floors},
\textit{dancing},
\textit{opening the door}, and
\textit{waving hands}. All personal data is collected with an IRB approval.
We generate 3D poses as pseudo labels using the egocentric and external images, which will be elaborated later. 
In terms of size, our EgoPW dataset is larger than existing in-the-wild 3D pose estimation datasets, like 3DPW~\cite{von2018recovering}, and has similar scale to the existing synthetic egocentric datasets, including the  Mo$^2$Cap$^2$~\cite{DBLP:journals/tvcg/XuCZRFST19} and the $x$R-egopose~\cite{DBLP:conf/iccv/TomePAB19} datasets. 

\subsection{Optimization for Generating Pseudo Labels}\label{subsec:optimizer}

In this section, we present an optimization method based on \cite{wang2021estimating} to generate pseudo labels for EgoPW.
Given a sequence, we split it into segments containing $B$ consecutive frames. 
For the egocentric frames $I^{ego}_{seq}$, we estimate the 3D poses represented by 15 joint locations in the coordinate system of the egocentric camera (called ``egocentric poses")  $\widetilde{\mathcal{P}}^{ego}_{seq}=\{\widetilde{\mathcal{P}}^{ego}_1,\dots,\widetilde{\mathcal{P}}^{ego}_B \}$, $\widetilde{\mathcal{P}}^{ego}_i \in \mathbb{R}^{15{\times}3}$, and 2D heatmaps $H^{ego}_{seq}=\{H^{ego}_1,\dots,H^{ego}_B \}$ using the Mo$^2$Cap$^2$ method \cite{DBLP:journals/tvcg/XuCZRFST19}.
Aside from egocentric poses, we also estimate the transformation matrix between the egocentric camera poses of two adjacent frames $[R_{seq}^{SLAM} \mid t_{seq}^{SLAM}] = \{[R_1^2 \mid t_1^2], \dots, [R_{B-1}^{B} \mid t_{B-1}^B]\}$ using ORB-SLAM2~\cite{murTRO2015}.
For the external frames $I^{ext}_{seq}$, we estimate the 3D poses (called ``external poses") $\mathcal{P}^{ext}_{seq}=\{\mathcal{P}^{ext}_1,\dots,\mathcal{P}^{ext}_B \}$, $\mathcal{P}^{ext}_i \in \mathbb{R}^{15 \times 3}$ using VIBE~\cite{DBLP:conf/cvpr/KocabasAB20} and 2D joints $\mathcal{J}^{ext}_{seq}=\{\mathcal{J}^{ext}_1,\dots,\mathcal{J}^{ext}_B \}$, $\mathcal{J}^{ext}_i \in \mathbb{R}^{15 \times 2}$ using openpose~\cite{cao2017realtime}. 

Next, following \cite{wang2021estimating}, we learn a latent space to encode an egocentric motion prior with a sequential VAE which consists of a CNN-based encoder $f_{enc}$ and decoder $f_{dec}$. We then optimize the egocentric pose by finding a latent vector $z$ such that the corresponding pose sequence $P^{ego}_{seq} = f_{dec}(z)$ minimizes the objective function:

\begin{equation}
\begin{aligned}
	E (\mathcal{P}_{seq}^{ego}, R_{seq}, t_{seq}) &= \lambda_R^{ego} E_{R}^{ego} + \lambda_R^{ext} E_{R}^{ext} + \lambda_J^{ego} E_{J}^{ego} \\
	& +  \lambda_J^{ext} E_{J}^{ext} + \lambda_T E_{T} + \lambda_B E_{B}\\
	& + \lambda_C E_{C} + \lambda_M E_{M}.
	\label{eq:optim}
\end{aligned}
\end{equation}
In this objective function, $E_{R}^{ego}$, $E_{J}^{ego}$,$E_{T}$, and $E_{B}$ are egocentric reprojection term, egocentric pose regularization term, motion smoothness regularization term and bone length regularization term, which are the same as those defined in~\cite{wang2021estimating}.
$E_{R}^{ext}$, $E_{J}^{ext}$, $E_{C}$, and $E_{M}$ are the external reprojection term, external 3D body pose regularization term, camera pose consistency term, and camera matrix regularization term, which will be described later. 
Please see the supplemental material for a detailed definition of each term.

Note that since the relative pose between external camera and egocentric camera is unknown, we also need to optimize the relative egocentric camera pose with respect to the external camera pose for each frame, i.e. the rotations  $R_{seq} = {R_1,\dots,R_B}$ and translations $t_{seq} = {t_1,\dots,t_B}$.

\vspace{-0.2em}
\paragraph{External Reprojection Term.}
In order to supervise the optimization process with the external 2D pose, we designed the external reprojection term which minimizes the difference between the projected 3D pose with the external 2D joints.
The energy term is defined as:
\begin{equation}
    E_{R}^{ext}(\mathcal{P}_{seq}^{ego}, R_{seq}, t_{seq}) = \sum_{i=1}^{B}\left\Vert \mathcal{J}_{i}^{ext} - K\left[R_i\mid t_i\right]\mathcal{P}_{i}^{ego} \right\Vert_2^2,
    \label{eq:ext_reproj}
\end{equation}
where $K$ is the intrinsic matrix of the external camera; $\left[R_{i} \mid t_{i} \right]$ is the pose of the egocentric camera in the $i$ th frame
w.r.t the external camera position.
In Eq.~\ref{eq:ext_reproj}, we first project the egocentric body pose $\mathcal{P}_{i}^{ego}$ to the 2D body pose in the external view with the egocentric camera pose $\left[R_{i} \mid t_{i} \right]$ and the intrinsic matrix $K$, and then compare the projected body pose with the 2D joints estimated by the openpose~\cite{cao2017realtime}. Since the relative pose between the external camera and egocentric camera are unknown at the beginning of the optimization, we optimize the egocentric camera pose $\left[R_{i} \mid t_{i} \right]$ simultaneously while optimizing the egocentric body pose $\mathcal{P}_{seq}^{ego}$.
In order to make the optimization process converge faster, we initialize the egocentric camera pose $\left[R_{i} \mid t_{i} \right]$ with the Perspective-n-Point algorithm~\cite{Gao2003CompleteSC}.

\vspace{-0.2em}
\paragraph{Camera Pose Consistency.}
We cannot get the accurate 3D pose only with the external reprojection term because the egocentric camera pose and the optimized body pose can be arbitrarily changed without violating the external reprojection constraint.
To alleviate this ambiguity, we introduce the camera consistency term $E_C$ as follows: 
\begin{equation}
\begin{aligned}
    E_{C}(R_{seq}, t_{seq}) & = \sum_{i=1}^{B-1}\left\Vert \begin{bmatrix} R_i  & t_i \\ 0 & 1 \end{bmatrix} \begin{bmatrix} R^{i+1}_i  & t^{i+1}_i \\ 0 & 1 \end{bmatrix} \right.\\
    & - \left. \begin{bmatrix} R_{i+1}  & t_{i+1} \\ 0 & 1 \end{bmatrix}\right\Vert_2,
    \label{eq:consistency}
\end{aligned}
\end{equation}
It enforces the egocentric camera pose at $(i + 1)$ th frame $[R_{i + 1} \mid t_{i + 1}]$ to be consistent with the pose obtained by transforming the egocentric camera pose at the $i$ th frame $[R_{i} \mid t_{i}]$ with the relative pose between the $i$ th and $(i + 1)$ th frame.

\vspace{-0.2em}
\paragraph{External 3D Body Pose Regularization.}
Besides the external reprojection term, we also use the external 3D body poses to supervise the optimization of the egocentric 3D body pose. 
We define the external 3D pose term which measures the difference between the external and the egocentric body poses after a rigid alignment:
\begin{equation}
	E_{J}(\mathcal{P}_{seq}^{ego}, \mathcal{P}_{seq}^{ext}) = \sum_{i=1}^{B}\left\Vert \mathcal{P}_{i}^{ext} - \left[R_{i}^{pa}\mid t_{i}^{pa}\right]\mathcal{P}_{i}^{ego} \right\Vert_2^2,
	\label{eq:ext_3d_pose}
\end{equation}
where $\left[R_{i}^{pa} \mid t_{i}^{pa} \right]$ is the transformation matrix calculated with Procrustes analysis, which rigidly aligns the external 3D pose estimation $\mathcal{P}_{i}^{ext}$ and the egocentric 3D pose $\mathcal{P}_{i}^{ego}$.

\begin{figure}
\centering
\includegraphics[width=0.98\linewidth]{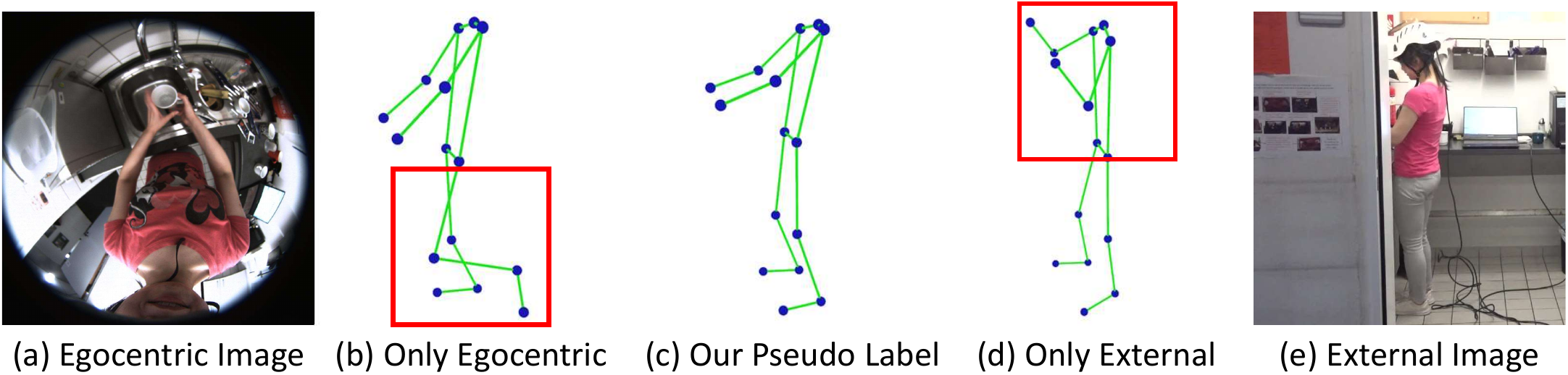}
\caption{Our pseudo label generation method combines the information from both the egocentric view and external view, therefore leading to more accurate pseudo labels (c). Only with the egocentric camera, the feet cannot be observed and well-tracked (b). Only with the external camera, the hands are occluded and result in the wrong result on the hand part (d).
}
\label{fig:optimization_example}
\end{figure}

By combining the body poses estimated from the egocentric view and external view, we can reconstruct more accurate pseudo labels.  
As shown in Fig.~\ref{fig:optimization_example}, the hands of the person are occluded in the external view, resulting in the tracking of the hands failing in the external view (Fig.~\ref{fig:optimization_example}, b), however, the hands can be clearly seen and tracked in the egocentric view (Fig.~\ref{fig:optimization_example}, d); 
on the other hand, the feet cannot be observed in the egocentric view and thus fail to be tracked in this view (Fig.~\ref{fig:optimization_example}, b), but can be easily viewed and tracked in the external view (Fig.~\ref{fig:optimization_example}, d). 
By joining the information from both views, we can successfully predict accurate 3D poses as the pseudo labels (Fig.~\ref{fig:optimization_example}, c).  
We note that the external camera is only used for generating the pseudo labels but at test time, only the egocentric camera is used. 

\paragraph{Camera Matrix Regularization.}
We constrain the camera rotation matrix $R_{i}$ to be orthogonal:

\begin{equation}
	E_{J}(R_{seq}) = \sum_{i=1}^{B}\left\Vert R_{i}^TR_{i} - I \right\Vert_2^2 \label{eq:matrix}.
\end{equation}

Different from previous single-view pose estimation methods which leverages the weak supervision from multiple views~\cite{iqbal2020weakly,rhodin2018learning,wandt2021canonpose,kocabas2019self}, our spatio-temporal optimization method generates the pseudo labels under the guidance of learned motion prior, making it robust to noisy and inaccurate 2D pose estimations which is common for the 2D pose estimation results from the egocentric view.

\subsection{Training Egocentric Pose Estimation Network}\label{subsec:train}

Through the optimization framework in Sec.~\ref{subsec:optimizer}, we can get accurate 3D pose pseudo labels $\mathcal{P}_{seq}^{ego}$ for each egocentric frame in the EgoPW dataset, which is further processed into the 2D heatmap $H_E$ and the distance between joints and egocentric camera $D_E$ with the fisheye camera model \cite{scaramuzza2014omnidirectional} described in supplementary materials. 

Afterward, we train a single-image based egocentric pose estimation network on both the synthetic dataset from Mo$^2$Cap$^2$ and the EgoPW dataset, as shown in the right part of Fig.~\ref{fig:framework}. The pose estimation network contains a feature extractor $\Theta$ which encodes an image into a feature vector and a pose estimator $\Psi$ which decodes the feature vector to 2D heatmaps and a distance vector. The 3D pose can be reconstructed from them with the fisheye camera model. Here, we note the synthetic dataset $S = \{I_S, H_S, D_S\}$ including synthetic images $I_S$ along with their corresponding heatmaps $H_S$ and distance labels $D_S$ from Mo$^2$Cap$^2$ dataset, and the EgoPW dataset $E = \{I_E^{ego}, H_E, D_E, I_E^{ext} \}$ including egocentric in-the-wild images $I_E^{ego}$ along with pseudo heatmaps $H_E$, distance labels $D_E$ and corresponding external images $I_E^{ext}$. 
During the training process, we train the egocentric pose estimation network with two reconstruction loss terms and two adversarial loss terms. The reconstruction losses are defined as the mean squared error (MSE) between the predicted heatmaps/distances  and heatmaps/distances from labels:

\begin{equation}
    \begin{aligned}
        L_{S} &= \text{mse}(\hat{H}_{S}, H_{S}) + \text{mse}(\hat{D}_{S}, D_{S})\\
        L_{E} &= \text{mse}(\hat{H}_{E}, H_{E}) + \text{mse}(\hat{D}_{E}, D_{E}),
    \end{aligned}
\end{equation}
where 
\begin{equation}
    \begin{aligned}
        \hat{H}_{S}, \hat{D}_{S} = \Psi(F_{S}),& F_{S} = \Theta(I_{S});\\
        \hat{H}_{E}, \hat{D}_{E} = \Psi(F_{E}^{ego}),& F_{E}^{ego} = \Theta(I_{E}^{ego}).
    \end{aligned}
\end{equation}

Two adversarial losses are separately designed for learning egocentric feature representation and bridging the domain gap between synthetic and real datasets. These two losses are described as follows.


\subsubsection{Adversarial Domain Adaptation}\label{subsubsec:da}
To bridge the domain gap between the synthetic and real data domains, following Tzeng~\etal~\cite{tzeng2017adversarial}, we introduce an adversarial discriminator $\Gamma$ which takes as input the feature vectors extracted from a synthetic image and an in-the-wild image, and determines if the feature is extracted from an in-the-wild image. 
The adversarial discriminator $\Gamma$ is trained with a cross-entropy loss:

\begin{equation}
    \mathcal{L}_{D} = -E[\log(\Gamma(\mathcal{F_S}))] - E[\log(1 - \Gamma(\mathcal{F_T}))].
\end{equation}

Once the discriminator $\Gamma$ has been trained, the feature extractor $\Theta$ maps the images from different domains to the same feature space such that the classifier $\Gamma$ cannot tell if the features are extracted from synthetic images or real images. Therefore, the pose estimator $\Psi$ can predict more accurate poses for the in-the-wild data. 


\subsubsection{Supervising Egocentric Feature Representation with External View}\label{subsubsec:ext_adv}

Although our new training dataset is large, the variation of identities in the dataset is still relatively limited (20 identities) compared with the existing large-scale external-view human datasets (thousands of identities). Generally speaking, the representations learned with these external-view datasets are of higher quality due to the large diversity of the datasets. To further improve the generalizability of our network and prevent overfitting to the training identities, we propose to supervise our egocentric representation by leveraging the high-quality third-person-view features.
From a transfer learning perspective, although following Mo$^2$Cap$^2$~\cite{DBLP:journals/tvcg/XuCZRFST19}, our egocentric network is pretrained on the third-person-view datasets, it can easily ``forget'' the learned knowledge while being finetuned on the synthetic dataset. The supervision from third-person-view features can prevent the egocentric features from deviating too much from those learned from large-scale real human images.

%
%
However, directly minimizing the distance between egocentric features $F_{E}^{ego}$ and external features $F_{E}^{ext}$ will not enhance the performance since the intermediate features of the egocentric and external view should be different from each other due to significant difference on the view direction and camera distortions.
%
%
To tackle this issue, we use the adversarial training strategy to align the feature representation from egocentric and external networks. Specifically, we use an adversarial discriminator $\Lambda$ which takes the feature vectors extracted from an egocentric image and the corresponding in-the-wild images and predicts if the feature is from egocentric or external images. The adversarial discriminator $\Lambda$ is trained with a cross-entropy loss:
\begin{equation}
    L_{V} = -E[\log(\Lambda(F_E^{ego}))] - E[\log(1 - \Lambda(F_E^{ext}))],
\end{equation}
where $F_{E}^{ext} = \Theta^{ext}(I_{E}^{ext})$ and $\Theta^{ext}$ is the feature extractor of external pose estimation network that shares exactly the same architecture as the egocentric pose estimation network. The parameters of the features extractor $\Theta^{ext}$ and the pose estimator $\Psi^{ext}$ of the external pose estimation network are obtained from the pretrained model in Xiao~\etal's work~\cite{xiao2018simple} and keep fixed during the training process.

Note that the deep layers of the pose estimation network usually represent the global semantic information of the human body~\cite{chu2017multi}, we use the output feature of the 4th res-block of ResNet-50 network~\cite{he2016deep} as the input to the discriminator $\Lambda$. Furthermore, the spatial position of the joints is quite different in the egocentric view and the external view, which will make the discriminator $\Lambda$ easily learn the difference between egocentric and external features. To solve this, we use an average pooling layer in the discriminator $\Lambda$ to spatially aggregate features, thus further eliminating the influence of spatial distribution between egocentric and external images. Please refer to the suppl. mat. for further details.

\begin{figure}
\centering
\includegraphics[width=0.98\linewidth]{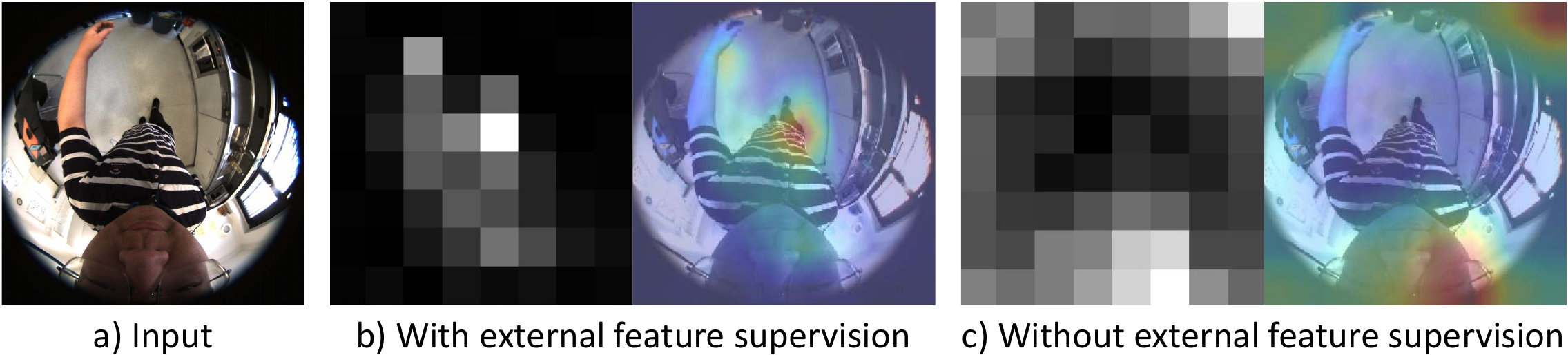}
\caption{The visualization of features with (b) or without (c) the adversarial supervision from external features. By supervising the training of the egocentric network with the feature representation from an external view, the egocentric network is able to focus on extracting the semantic features of the human body.
}
\label{fig:feature}
\end{figure}

During the training process, the egocentric pose estimation network is trained to produce the features $F_E^{ego}$ to fool the domain classifier $\Lambda$ such that it cannot distinguish whether the feature is from an egocentric or external image.
%
To achieve this, the egocentric network learns to pay more attention to the relevant parts of the input image,~\ie, the human body, which is demonstrated in Fig.~\ref{fig:feature}. 

\section{Experiments}

\subsection{Datasets}

We quantitatively evaluate our finetuned network on the real-world dataset from Mo$^2$Cap$^2$ \cite{DBLP:journals/tvcg/XuCZRFST19} and Wang \etal~\cite{wang2021estimating}. The real-world dataset in Mo$^2$Cap$^2$~\cite{DBLP:journals/tvcg/XuCZRFST19} contains 2.7k frames of two people captured in indoor and outdoor scenes, and that in Wang \etal~\cite{wang2021estimating} contains 12k frames of two people captured in the studio. To measure the accuracy of our pseudo labels, we evaluate our optimization method (Sec.~\ref{subsec:optimizer}) only on the dataset from Wang \etal~\cite{wang2021estimating} since the Mo$^2$Cap$^2$ dataset does not include the external view.

To evaluate our method on the in-the-wild data, we also conduct a qualitative evaluation on the test set of the EgoPW dataset. The EgoPW dataset will be made publicly available, and more details and comparisons to other datasets are included in the supplementary materials.

\subsection{Evaluation Metrics}
We measure the results of our method as well as other baseline methods with two metrics, PA-MPJPE and BA-MPJPE, which estimate the accuracy of a single body pose. For \textbf{PA-MPJPE}, we rigidly align the estimated pose $\mathcal{\hat{P}}$ of each frame to the ground truth pose $\mathcal{P}$ using Procrustes analysis \cite{kendall1989survey}. In order to eliminate the influence of the body scale, we also report the \textbf{BA-MPJPE} scores. In this metric,  we first resize the bone length of each predicted body pose $\mathcal{\hat{P}}$ and ground truth body pose $\mathcal{P}$ to the bone length of a standard skeleton. Then, we calculate the PA-MPJPE between the two resulting poses. 

\begin{figure*}
\centering
\includegraphics[width=0.975\linewidth]{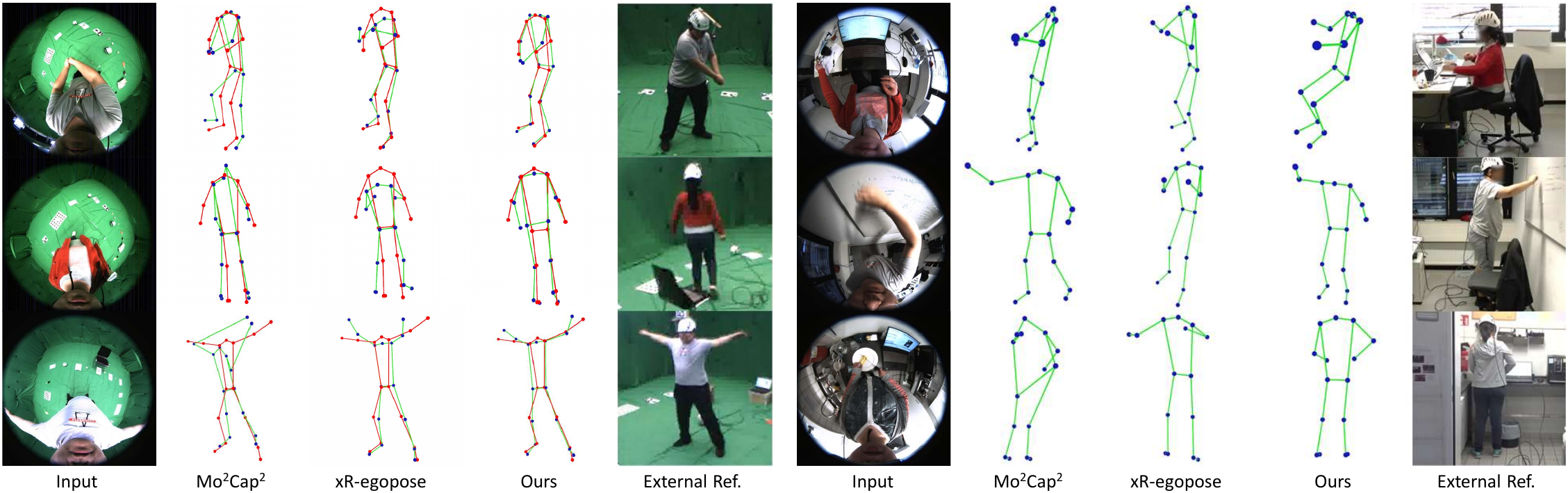}
\caption{Qualitative comparison between our method and the state-of-the-art methods. From left to right: input image, Mo$^2$Cap$^2$ result, $x$R-egopose result, our result, and external image. The ground truth pose is shown in red. Note that the external images are not used during inference.}
\label{fig:results}
\end{figure*}

\subsection{Pseudo Label Generation}\label{subsec:pseudo_labels}

\begin{table}[h]
\begin{center}
\small
\begin{tabular}{p{0.19\textwidth}
>{\centering}p{0.1\textwidth}
>{\centering\arraybackslash}p{0.1\textwidth}}
\hlineB{2.5}
Method & PA-MPJPE & BA-MPJPE \\ \hline
Mo$^2$Cap$^2$ & 102.3 & 74.46 \\
$x$R-egopose & 112.0 & 87.20 \\
Wang \etal~\cite{wang2021estimating} & 83.40 & 63.88 \\
VIBE~\cite{DBLP:conf/cvpr/KocabasAB20} & 68.13 & 52.99 \\
\hline
Our Optimizer & \textbf{57.19} & \textbf{46.14}\\
\hlineB{2.5}
\end{tabular}
\end{center}
\caption{The performance of our optimization framework on  Wang \etal's dataset. Our optimization method outperforms all previous egocentric pose estimation methods and the state-of-the-art pose estimation method of the external view.}
\label{table:pseudo_labels}
\end{table}

In this paper, we first generate the pseudo labels with the optimization framework (Sec.~\ref{subsec:optimizer}) and use them to train our network. Thus, pseudo labels with higher accuracy generally lead to better network performance.
In this experiment, we evaluate the accuracy of pseudo labels on Wang~\etal's dataset and show the results in Table~\ref{table:pseudo_labels}. 
This table shows that our method outperforms all the baseline methods by leveraging both the egocentric view and external view during optimization. Note that though compared in Table~\ref{table:pseudo_labels}, we cannot use any external-view based pose estimation method, \eg VIBE~\cite{DBLP:conf/cvpr/KocabasAB20} and 3DPW~\cite{von2018recovering}, for training the egocentric pose estimation network. This is because the relative pose between the external and egocentric camera is unknown, making it impossible to obtain the egocentric body pose only from the external view. Compared with our optimization approach, the method in~\cite{wang2021estimating} performs worse due to the lack of external-view supervision. 

\subsection{Comparisons on 3D Pose Estimation}

\begin{table}[h]
\begin{center}
\small
\begin{tabularx}{0.47\textwidth} { 
   >{\raggedright\arraybackslash}X 
   >{\centering\arraybackslash}X 
   >{\centering\arraybackslash}X  }
\hlineB{2.5}
Method & PA-MPJPE & BA-MPJPE \\
\hline
\multicolumn{2}{l}{\textbf{Wang \etal's test dataset}} \\
\hline
Rhodin \etal~\cite{rhodin2018learning} & 89.67 & 73.56 \\
Mo$^2$Cap$^2$~\cite{DBLP:journals/tvcg/XuCZRFST19} & 102.3 & 74.46 \\
$x$R-egopose~\cite{DBLP:conf/iccv/TomePAB19} & 112.0 & 87.20 \\
Ours &\textbf{81.71} & \textbf{64.87} \\
\hlineB{2.5}
\multicolumn{2}{l}{\textbf{Mo$^2$Cap$^2$ test dataset}} \\
\hline
Rhodin \etal~\cite{rhodin2018learning} & 97.69 & 76.92 \\
Mo$^2$Cap$^2$~\cite{DBLP:journals/tvcg/XuCZRFST19} & 91.16 & 70.75 \\
$x$R-egopose~\cite{DBLP:conf/iccv/TomePAB19} & 86.85 & 66.54 \\
Ours & \textbf{83.17} & \textbf{64.33} \\
\hlineB{2.5}
\end{tabularx}
\end{center}
\caption{The experimental results on Wang~\etal's test dataset and Mo$^2$Cap$^2$ test dataset \cite{DBLP:journals/tvcg/XuCZRFST19}. Our method outperforms the state-of-the-art methods, i.e. Mo$^2$Cap$^2$ \cite{DBLP:journals/tvcg/XuCZRFST19} and $x$R-egopose \cite{DBLP:conf/iccv/TomePAB19} on both of the metrics. Our method also outperforms a video-based method~\cite{wang2021estimating} in terms of  PA-MPJPE.
}
\label{table:main}
\end{table}

We compare our approach with previous single-frame-based methods on the test dataset from~\cite{wang2021estimating} under the ``Wang~\etal's test dataset'' in Table~\ref{table:main}. Since the code or the predictions of $x$R-egopose are not publicly available, we use our reimplementation of $x$R-egopose instead. On this dataset, our method outperforms Mo$^2$Cap$^2$ by 20.1\% and $x$R-egopose by 27.0\% respectively. 
We also compared with previous methods on the Mo$^2$Cap$^2$ test dataset and show the results under the ``Mo$^2$Cap$^2$ test dataset'' in Table~\ref{table:main}. On the Mo$^2$Cap$^2$ test dataset, our method performs better than Mo$^2$Cap$^2$ and $x$R-egopose by 8.8\% and 4.2\%, respectively.


From the results in Table~\ref{table:main}, we can see that our approach outperforms all previous methods on the single-frame egocentric pose estimation task. More quantitative results on each type of motion are available in the supplementary material. For the qualitative comparison, we show the results of our method on the studio dataset and in-the-wild dataset in Fig.~\ref{fig:results}. Our method performs much better compared with Mo$^2$Cap$^2$ and $x$R-egopose, especially for the in-the-wild cases where the body parts are occluded. Please refer to the supplementary materials for more qualitative results.

We also compared our method with Rhodin~\etal's method~\cite{rhodin2018learning}, which leverages the weak supervision from multiple views to supervise the training of a single view pose estimation network. In our EgoPW dataset, we only have one egocentric and one external view. Thus, we fix the 3D pose estimation network for the external view and only train the egocentric pose estimation network. Following Rhodin~\etal~\cite{rhodin2018learning}, we align the prediction from the egocentric and external view with Procrustes analysis and calculate the loss proposed by Rhodin~\etal. Our result in Table~\ref{table:main} shows that the performance of our method is better. This is mainly because our spatio-temporal optimization method predicts accurate and temporally stable 3D poses as pseudo labels, while other methods suffer seriously from inaccurate egocentric pose estimations.

\subsection{Ablation Study}\label{subsec:ablation}

\begin{table}[h]
\begin{center}
\small
\begin{tabular}{p{0.2\textwidth}
>{\centering}p{0.1\textwidth}
>{\centering\arraybackslash}p{0.1\textwidth}}
\hlineB{2.5}
Method & PA-MPJPE & BA-MPJPE \\ 
\hline
w/o external view  & 90.05 & 68.99 \\
w/o learning representation  & 85.46 & 67.01 \\
w/o domain adaptation  & 84.22 & 66.48 \\
Unsupervised DA  & 91.56 & 69.17 \\
\hline
Ours & \textbf{81.71} & \textbf{64.87}\\
\hlineB{2.5}
\end{tabular}
\end{center}
\caption{The quantitative results of ablation study.}
\label{table:ablation}
\end{table}

\begin{figure}
\centering
\includegraphics[width=0.98\linewidth]{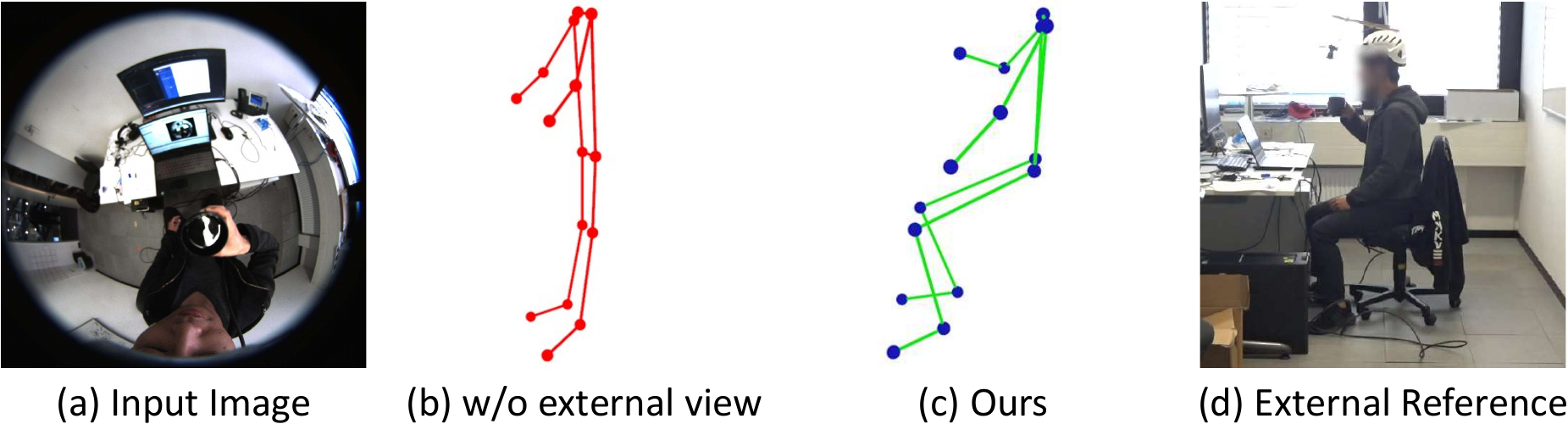}
\caption{The results of our method with (c) and without external view (b). The network cannot predict accurate poses for the occluded cases without the external view supervision. 
The external view is only for visualization and not used for predicting the pose.
}
\label{fig:wo_external}
\vspace{-0.2em}
\end{figure}

\paragraph{Supervision from the external view.}
In our work, we introduce the external view as supervision for training the network on the real EgoPW dataset. The external view enables generating accurate pseudo labels, especially when the human body parts are occluded in the egocentric view but can be observed in the external view. Without the external view, the obtained pseudo labels are less accurate and will further affect the network performance. In order to demonstrate this, 
we firstly generate the 3D poses as pseudo labels with Wang~\etal's method, \ie without any external supervision, and then train the pose estimation network on these new pseudo labels.
The result is shown in the ``w/o external view'' row of Table~\ref{table:ablation}. We also show the qualitative results with and without external-view supervision in Fig.~\ref{fig:wo_external}. Both the qualitative and quantitative results demonstrate that with the external supervision, the performance of our pose estimation network is significantly better especially on occluded cases.

\paragraph{Learning egocentric feature representation and bridging the domain gap with adversarial training.} 
In our work, we train the pose estimation network with two adversarial components in order to learn the feature representation of the egocentric human body (Sec.~\ref{subsubsec:ext_adv}) and bridge the domain gap between synthetic and real images (Sec.~\ref{subsubsec:da}). In order to demonstrate the effectiveness of both modules, we removed the domain classifier $\Lambda$ in our training process and show the results in the row of ``w/o learning representation'' in Table~\ref{table:ablation}. We also removed the domain classifier $\Gamma$, train the network without $L_{D}$ and show the quantitative results in the row of ``w/o domain adaptation'' in Table~\ref{table:ablation}. After moving any of the two components, our method suffers from the performance drop, which demonstrates the effectiveness of both the feature representation learning module and the domain adaptation module.

\paragraph{Comparison with only using unsupervised domain adaptation.}
In this experiment, we compare our approach with the unsupervised adversarial domain adaptation method~\cite{tzeng2017adversarial} which is commonly used for transfer learning tasks. We train the network only with the $L_S$ and $L_D$ in the adversarial domain adaptation module (Sec.~\ref{subsubsec:da}) and show the results in the ``Unsupervised DA'' of the Table~\ref{table:ablation}. Our approach outperforms the unsupervised domain adaptation method due to our high-quality pseudo labels.

\section{Conclusions}
In this paper, we have proposed a new approach to egocentric human pose estimation with a single head-mounted fisheye camera. We collected a new in-the-wild egocentric dataset (EgoPW) and designed a new optimization method to generate accurate egocentric poses as pseudo labels. Next, we supervise the egocentric pose estimation network with the pseudo labels and the features from the external network.
The experiments show that our method outperforms all of the state-of-the-art methods both qualitatively and quantitatively and our method also works well under severe occlusion. 
As future work, we would like to develop a video-based method for estimating temporally-consistent egocentric poses from an in-the-wild video. 

\noindent\textbf{Limitations.} The accuracy of pseudo labels in our method is constrained by our in-the-wild capture system, which only contains one egocentric view and one external view, and further constrains the performance of our method. To solve this, we can fuse different sensors, including IMUs and depth cameras, for capturing the in-the-wild dataset, which can be explored in our follow-up work.

{\small
\bibliographystyle{ieee_fullname}
\bibliography{egbib}
}

\end{document}